\documentclass[11pt,a4paper]{article}
\pdfoutput=1
\usepackage{naaclhlt2019}
\usepackage{times}
\usepackage{latexsym}

\usepackage{url}

\PassOptionsToPackage{usenames,dvipsnames,table}{xcolor}
\usepackage[utf8]{inputenc} 
\usepackage[T1]{fontenc}    
\usepackage{booktabs}       
\usepackage{amsfonts}       
\usepackage{nicefrac}       
\usepackage{microtype}      
\usepackage{amssymb}
\usepackage{amsmath}
\usepackage{multirow}
\usepackage{amsfonts}
\usepackage{booktabs}
\usepackage{rotating}
\usepackage{subfig}
\usepackage{array}
\usepackage{booktabs}
\usepackage{amsfonts}
\usepackage{enumitem}
\usepackage{makecell}
\usepackage{xcolor}
\usepackage[toc,page]{appendix}
\usepackage{amsmath}
\usepackage{xspace}
\usepackage{mdwlist}
\usepackage{graphics}
\usepackage{color}
\usepackage{rotating}
\usepackage{booktabs}
\usepackage{epsfig}
\usepackage{alltt}
\usepackage{moreverb}
\usepackage{fancyvrb}
\usepackage{enumerate}
\usepackage{colortbl}      
\usepackage{xspace}
\usepackage{relsize}
\usepackage{array}
\usepackage{amsmath}
\usepackage{amsfonts}
\usepackage{txfonts}
\usepackage{caption}
\DeclareCaptionFont{10pt}{\fontsize{10pt}{12pt}\selectfont}
\usepackage{rotating}
\usepackage{latexsym}
\usepackage{multirow}
\usepackage{booktabs}
\usepackage{float}
\usepackage{array,etoolbox}
\usepackage{enumitem}
\usepackage{subfig}
\usepackage{setspace}
\usepackage{float}
\usepackage{mathtools}
\usepackage{bm}
\usepackage{xcolor}
\usepackage[toc,page]{appendix}
\usepackage{soul}
\usepackage[draft]{pgf}
\usepackage{makecell}
\usepackage[normalem]{ulem}
\usepackage{amssymb}
\usepackage{wasysym,textcomp}
\usepackage[toc,page]{appendix}
\usepackage{wrapfig}

\usepackage[symbol]{footmisc}

\usepackage{etoolbox}
\newtoggle{draft}
\toggletrue{draft}
\iftoggle{draft}{
\newcommand{\xinya}[1]{\textcolor{Blue}{[#1 \textsc{--Xinya}]}}
\newcommand{\peter}[1]{\textcolor{Blue}{[#1 \textsc{--Pete}]}}
\newcommand{\niket}[1]{\textcolor{Orange}{[#1 \textsc{--Niket}]}}
\newcommand{\bhavana}[1]{\textcolor{Green}{[#1 \textsc{--Bhavana}]}}
\newcommand{\todo}[1]{\textcolor{red}{[#1 \textsc{--TODO}]}}
}{
\newcommand{\xinya}[1]{}
\newcommand{\peter}[1]{}
\newcommand{\niket}[1]{}
\newcommand{\bhavana}[1]{}
\newcommand{\todo}[1]{}
}
\definecolor{Red}{rgb}{1,0,0}
\newcommand{\finaledit}[1]{\textcolor{black}{#1}}

\newcommand{\statechange}[1]{\texttt{\textit{#1}}}
\newcommand{\entity}[1]{\texttt{#1}}
\newcommand\prostruct{\textsc{ProStruct}}
\newcommand\proSSL{\textsc{LaCE}}
\newcommand\proSSLexpansion{Label Consistency Explorer}

\newcommand{\com}[1]{}

\newcommand{\eat}[1]{}
\mathchardef\mhyphen="2D
\newenvironment{ite}{                     
     \parskip 0cm \begin{itemize} \parskip 0cm \parsep 0cm \itemsep 0cm \topsep 0cm}{
        \end{itemize}} 

\makeatletter
\g@addto@macro\normalsize{%
  \setlength\abovedisplayskip{1pt}
  \setlength\belowdisplayskip{1pt}
  \setlength\abovedisplayshortskip{1pt}
  \setlength\belowdisplayshortskip{1pt}
}
\makeatother

\newcommand{\model}{\textsc{LaCE}}

\aclfinalcopy
\def\aclpaperid{1889} 

\title{Be Consistent! Improving Procedural Text Comprehension \\
using Label Consistency}
\author{Xinya Du{\normalfont \textsuperscript{1}\footnotemark}  \quad Bhavana Dalvi Mishra{\normalfont \textsuperscript{2}} \quad  Niket Tandon{\normalfont \textsuperscript{2}} \quad Antoine Bosselut{\normalfont \textsuperscript{2}} \\ 
        {\bf Wen-tau Yih{\normalfont \textsuperscript{2}}} \quad {\bf Peter Clark{\normalfont \textsuperscript{2}}} \quad {\bf Claire Cardie{\normalfont \textsuperscript{1}}} \\
        \textsuperscript{1}Department of Computer Science, Cornell University, Ithaca, NY\\ 
        {\tt \{xdu, cardie\}@cs.cornell.edu} \\
        \textsuperscript{2}Allen Institute for Artificial Intelligence, Seattle, WA \\
        {\tt \{bhavanad, nikett, antoineb, scottyih, peterc\}@allenai.org}
        }

\begin{document}

\maketitle

\begin{abstract}
Our goal is procedural text comprehension, \finaledit{namely} tracking how the properties of entities
(e.g., their location) change with time given a procedural text (e.g., a paragraph about photosynthesis, a recipe).
This task is challenging as the world is changing throughout the text, and
despite recent advances, current systems still struggle with this task.
Our approach is to leverage the fact that, for many procedural texts, multiple independent
descriptions are readily available, and that predictions from them should be consistent
(label consistency). We present a new learning framework that leverages
label consistency during training, allowing consistency bias to be built into
the model. Evaluation on a standard benchmark dataset for procedural text,
ProPara~\cite{propara-naacl18}, shows that our approach significantly improves prediction performance (F1) over
prior state-of-the-art systems.

\end{abstract}

\footnotetext[1]{ Work done while at the Allen Institute for Artificial Intelligence.}

\section{Introduction}

We address the task of procedural text comprehension, \finaledit{namely tracking
how the properties of entities (e.g., their location) change with time throughout
the procedure (e.g., photosynthesis, a cooking recipe).}
\eat{We address the task of procedural text comprehension, i.e., tracking how the properties of entities
(e.g., their location) change throughout the course of a procedural text
\finaledit{that} describes a process or procedure (e.g., photosynthesis or how to make sugar cookies). }
This ability is an important part of text understanding,
allowing the reader to infer unstated facts \finaledit{such as} how ingredients change during a recipe, what 
the inputs and outputs of a scientific process are, or who met whom in a news article about a political meeting. Although several procedural text comprehension systems have emerged recently 
(e.g., EntNet \cite{Henaff2016TrackingTW}, NPN \cite{npn}, and ProStruct \cite{propara-emnlp18}), they still make numerous prediction errors. A major challenge is
that fully annotated training data for this task is expensive to collect, because
many state changes by multiple entities may occur in a single text, requiring complex annotation.

To address this challenge, and thus improve performance,
our goals are two-fold: first, to better leverage the training data for procedural text
comprehension that {\it is}
available, and second, to utilize additional
unlabeled data for the task (semi-supervised learning).
Our approach in each case is to exploit {\it label consistency}, the
property that two distinct texts covering the same procedure should be generally consistent in terms of the state changes that they describe, which constitute the labels to be predicted for the text.
For example, in different texts describing photosynthesis, we expect them to
be generally consistent about what happens to oxygen (e.g., that it is created),
even if the wordings differ (Figure~\ref{example}).

\begin{figure}[t]
\fbox{%
  \parbox{0.44\textwidth}{%
(1) ...\textcolor{blue}{\bf oxygen} is given off... \\
(2) ...the plant produces \textcolor{blue}{\bf oxygen}... \\
(3) ...is used to create sugar and \textcolor{blue}{\bf oxygen}...
}}
\caption{
Fragments from three independent texts about photosynthesis. Although (1) is
ambiguous as to whether oxygen is being created or merely moved, evidence
from (2) and (3) suggests it is being created, helping to correctly interpret (1).
More generally, encouraging consistency between predictions from different
paragraphs about the same process/procedure can improve performance.}
\label{example}
\end{figure}

\begin{figure*}[!htb]
\begin{center}
{\includegraphics[width=\textwidth]{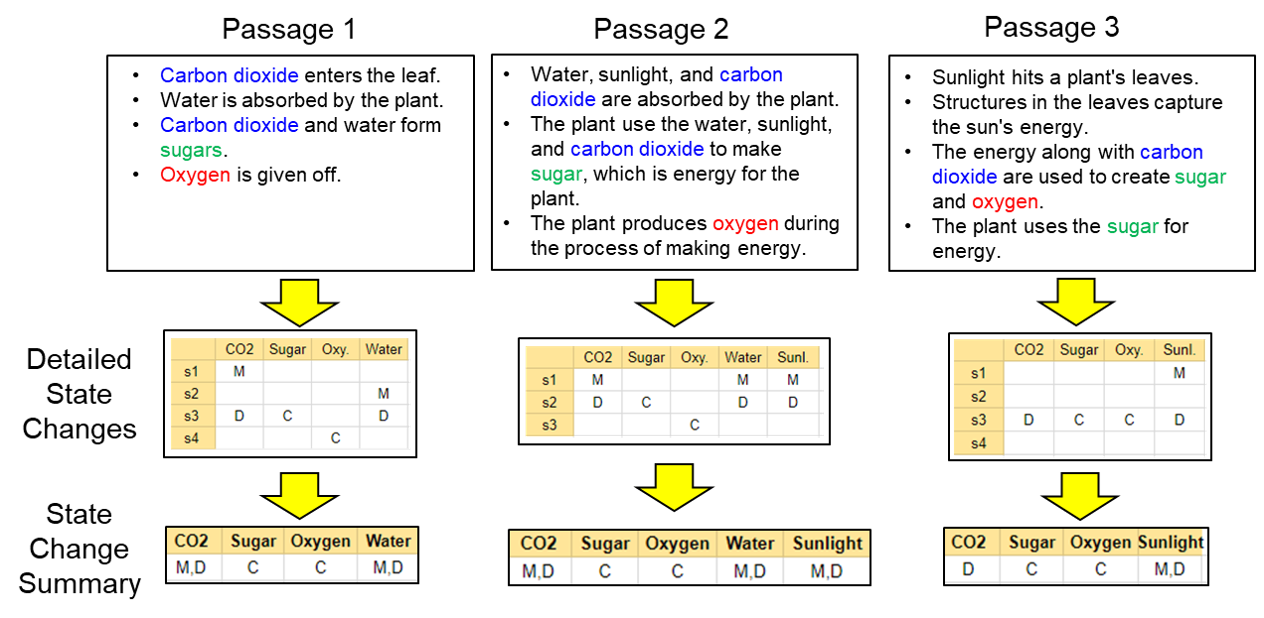}}
\end{center}
\caption{Three (simplified) passages from ProPara describing photosynthesis, the (gold) state changes each entity undergoes at each step \finaledit{$s_1, s_2, …, s_T$}, and the summary of state changes that each entity undergoes (an aggregation of the step-by-step changes), where \finaledit{\statechange{M = MOVED, D = DESTROYED, C = CREATED}}. Although the language and detailed changes for each passage differ considerably, the {\it \bf overall summaries are largely consistent} (e.g., sugar is \statechange{CREATED} in all three). We exploit this consistency when training a model to make these predictions, by biasing the model to prefer predictions whose summary is consistent with the (predicted) summaries of other passages about the same topic. Note that in the summary, we do not care about the order in which state changes happen, so summary \statechange{M, D} for participant \entity{CO$_2$} in passage 1 denotes a set of state changes rather than a sequence of state changes.}
\label{fig:example_multiple_paragraphs}
\end{figure*}

%

Using multiple, distinct passages to understand a process or procedure is challenging. Although 
the texts describe the same process, they might express the underlying facts at different levels of granularity, using different wordings, and including or omitting different details. As a result, the details may differ between paragraphs, making them hard to align and to check for consistency. Nonetheless, even if the details differ, we conjecture that the top-level {\it summaries} of each paragraph, which describe the types of state change that each entity undergoes, will be mostly consistent.
For example, although independent texts describing photosynthesis vary tremendously, we expect them to be consistent about what generally happens to sugar, e.g., that it is created (Figure~\ref{fig:example_multiple_paragraphs}).

In this paper, we introduce a new training framework, called \proSSL~(\proSSLexpansion), that leverages
label consistency among paragraph summaries.
In particular, it encourages label consistency during 
end-to-end training of a neural model, allowing consistency bias to improve the model itself, rather than be enforced in a
post-processing step, e.g., posterior regularization \cite{Ganchev2010PosteriorReg}. 
We evaluate on a standard benchmark for procedural text comprehension, called ProPara
\cite{propara-naacl18}. We show that this approach achieves a new state-of-the-art performance
in the fully supervised setting (when all paragraphs are annotated), and also demonstrate
that it improves performance in the
semi-supervised setting (using additional, unlabeled paragraphs)
with limited training data. In the latter case, summary
predictions from labeled data act as noisy gold labels for the unlabeled data,
allowing additional learning to occur. Our contributions are thus:
\begin{ite}
\item[1.] A new learning framework, \proSSL, applied to procedural text comprehension that improves the label consistency among different paragraphs on the same topic.
\item[2.] Experimental results demonstrating that \model~achieves state-of-the-art performance on a standard benchmark dataset, ProPara, for procedural text.
\end{ite}


\section{Related Work\label{related-work}}
Our work is related to several important branches of work in both NLP and ML, as we
now summarize.

\vspace{0.2cm} \noindent \textbf{Leveraging Label Consistency}
Leveraging information about label consistency (i.e., similar instances should have consistent labels at a certain granularity) is an effective idea. It has been studied in computer vision~\cite{haeusser2017learning, Chen2018Text2ShapeGS} and IR~\cite{clarke2001exploiting, Dumais2002WebQA}.
Learning by association~\cite{haeusser2017learning} establishes implicit cross-modal links between similar descriptions and leverage more unlabeled data during training. 
\newcite{Schtze2018TwoMF, hangya2018two} adapt the similar idea to exploit unlabeled data for the cross-lingual classification. 
\finaledit{We extend this line of research in two ways: by developing a framework allowing it to be applied to the task of structure prediction;
and by incorporating label consistency into the model itself via end-to-end training, rather than enforcing consistency as a post-processing step.}
\eat{
We extend this line of research by leveraging label consistency while training a neural model for a structured prediction task (i.e., summary of entity states change should be consistent among paragraphs with similar topic).}

\vspace{0.2cm} \noindent \textbf{Semi-supervised Learning Approaches}
Besides utilizing the label consistency knowledge, our learning framework is also able to use unlabeled paragraphs, which fits in the literature of semi-supervised learning approaches (for NLP). \newcite{Zhou2003LearningWL} propose an iterative label propagation algorithm similar to spectral clustering.
\finaledit{\newcite{zhu2003semi} propose a semi-supervised learning framework via harmonic energy minimization for data graph.} \newcite{Talukdar2008WeaklySupervisedAO} propose a graph-based semi-supervised label propagation algorithm for acquiring open-domain labeled classes and their instances from a combination of unstructured and structured text sources. 
\finaledit{Our framework extends these ideas by introducing the notion of groups (examples that are expected to be similar) and summaries (what similarities are expected), applied in an end-to-end-framework.}
\eat{
**PEC** This is (a) too much into technical details that haven't been discussed (b) unclear and unsubstantiated claims ("more balanced", "more efficient and exact") - not appropriate for a professional paper.

\finaledit{These graph based approaches optimize the joint loss combining supervised loss over training labels and label consistency among similar examples. Our framework is a variant of this approach by optimizing for such joint loss over smaller groups of data so that the two loss terms become more balanced and the propagation becomes more efficient and exact. We describe in Section~\ref{sec:problem-definition} that such groups occur naturally in procedural text and can admit a structural summary level as consistency loss.}
}
\eat{
**PEC** This seems irrelevant to what we're doing? 
Our work applies these ideas 
\finaledit{to structure prediction, using an end-to-end training framework.}
}

\vspace{0.2cm} \noindent \textbf{Procedural Text Understanding and Reading Comprehension}
There has been a growing interest in procedural text understanding/QA recently. The ProcessBank dataset~\cite{berant2014modeling} asks questions about event ordering and event arguments for biology processes. 
bAbI~\cite{weston2015towards} includes questions about movement of entities, however it's synthetically generated and with a small lexicon. \newcite{kiddon2015mise}'s \textsc{RECIPES} dataset introduces the task of predicting the locations of cooking ingredients, and \newcite{kiddon2016globally} for recipe generation. In this paper, we continue
this line of exploration using ProPara, and illustrate how the previous
two lines of work (label consistency and semi-supervised learning) can be integrated.

\section{Problem Definition}

\subsection{Input and Output}
\label{sec:problem-definition}

\begin{figure*}[ht]
\begin{center}
{\includegraphics[width=1.9\columnwidth]{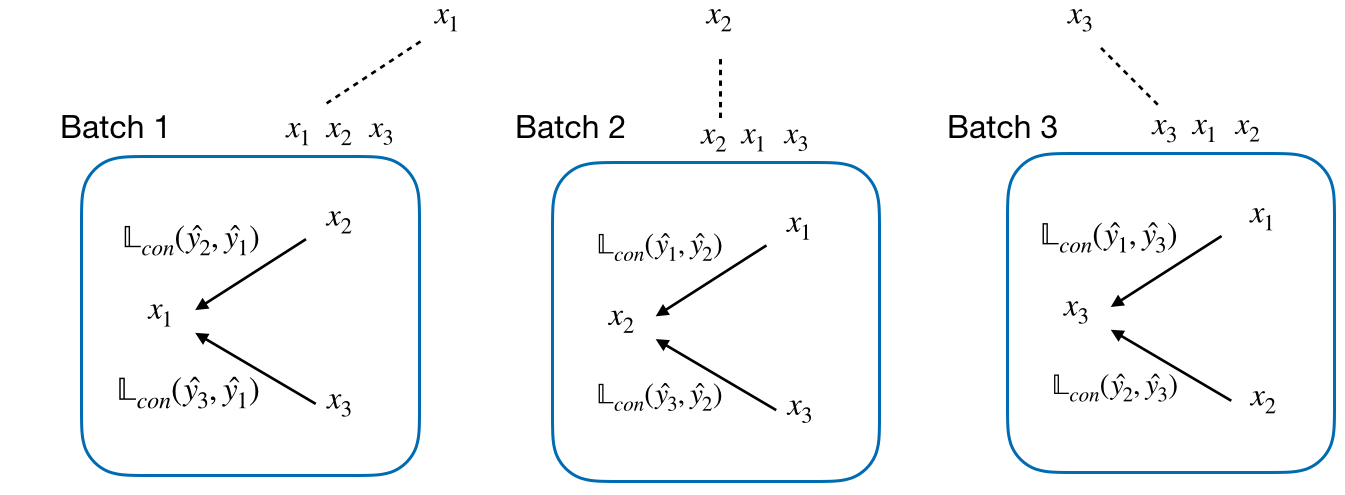}}
\end{center}
\caption{Example of batches constructed from a group (here, the group contains three labeled examples $x_1, x_2, x_3$). From three examples, three batches are constructed. \eat{We compute consistency loss (Equation \ref{eq:tau}) between the first element in the batch and the remaining elements.} \finaledit{Taking the predicted labels for the first element in the batch as reference we compute the consistency loss
for the remaining elements.}}
\label{fig:prossl_batching_only}
\end{figure*}

A general condition for applying our method is having multiple examples where, for {\it some} properties,
we expect to see similar values. For example, for procedural text, we expect paragraphs about
the same process to be similar in terms of which entities move, are created, and destroyed;
for different news stories about a political meeting, we expect top-level features (e.g.,
where the meeting took place, who attended) to be similar; for different recipes for the
same item, we expect loosely similar ingredients and steps; and for different images of the same
person, we expect some high-level characteristics (e.g., height, face shape) to be similar.
Note that this condition does not apply to every learning situation; it only applies
when training examples can be grouped, where all group members are expected to share some
characteristics that we can identify (besides the label used to form the groups in the first place).

More formally, for training, the input is a set of labeled examples $(x_{gi},y_{gi} )$
(where $y_{gi} $ are the labels for $x_{gi} $), partitioned into $G$ groups,
where the $g$ subscript denotes which group each example
belongs to. Groups are defined such that examples of the same group $g$ are expected to have
similar labels for a subset of labels $y_{gi}$. We call this subset the
{\bf summary labels}. We assume that both the groupings and the identity of the summary labels are provided.
The output of training is a model $M$ for labeling new examples. For testing, the input is the model $M$ and a set of
unlabeled (and ungrouped) examples $x_t$, and the output are their predicted labels $\hat{y_t}$.
Note that this formulation is agnostic to the learning algorithm used.
Later, we will consider both the fully supervised setting (all training examples
are labeled) and semi-supervised setting (only a subset are labeled).

\subsection{Instantiation}
\label{instantiation}
We instantiate this framework for procedural text comprehension, using the ProPara task  \cite{propara-naacl18}.
In this task, $x_{gi}$ are paragraphs of text describing a process (e.g., photosynthesis),
the labels $y_{gi}$ describe the state changes that each entity in the paragraph undergoes at
each step (sentence) (e.g., that oxygen is created in step 2),
and the groups are paragraphs about the same topic (ProPara tags each
paragraph with a topic, e.g., there are three paragraphs in ProPara
describing photosynthesis). More precisely, each $x_{gi}$ consists of:
\begin{ite}
\item the name (topic) of a process, e.g., photosynthesis
\item a sequence (paragraph) of sentences $S = [s_{1},...,s_{T}]$ that describes that process
\item the set of entities $E$ mentioned in that text, e.g., oxygen, sugar
\end{ite}
and the targets (labels) to predict are:
\begin{ite}
\item the state changes that each entity in $E$ undergoes at each step (sentence) of the process,
  where a state change is one of \{\statechange{Moved,Created,Destroyed,None}\}. These state changes can be
  conveniently expressed using a $|S| \times |E|$ matrix (Figure~\ref{fig:example_multiple_paragraphs}).
  State changes also include arguments, e.g., the source and destination of a move.
  We omit these in this paper to simplify the description.
\end{ite}
Finally, we define the summary labels as the set of state changes that
each entity undergoes at {\it some} point in the process, without
concern for when. For example, in Passage 1 in Figure~\ref{fig:example_multiple_paragraphs},
CO${_2}$ is \statechange{Moved} (M) and \statechange{Destroyed} (D), while sugar is \statechange{Created} (C).
These summary labels can be computed from the state-change matrix
by aggregating the state changes for each entity over all steps.
Our assumption here is that these summaries will generally be the
same (i.e., consistent) for different paragraphs about the same topic.
\model~then exploits this assumption by encouraging 
this inter-paragraph consistency during training,
as we now describe.

\section{Label Consistency Explorer: \model}

\subsection{The \model{} Learning Framework} \label{sec:lace_overview}


\begin{figure*}[t]
\begin{center}
{\includegraphics[width=2.0\columnwidth]{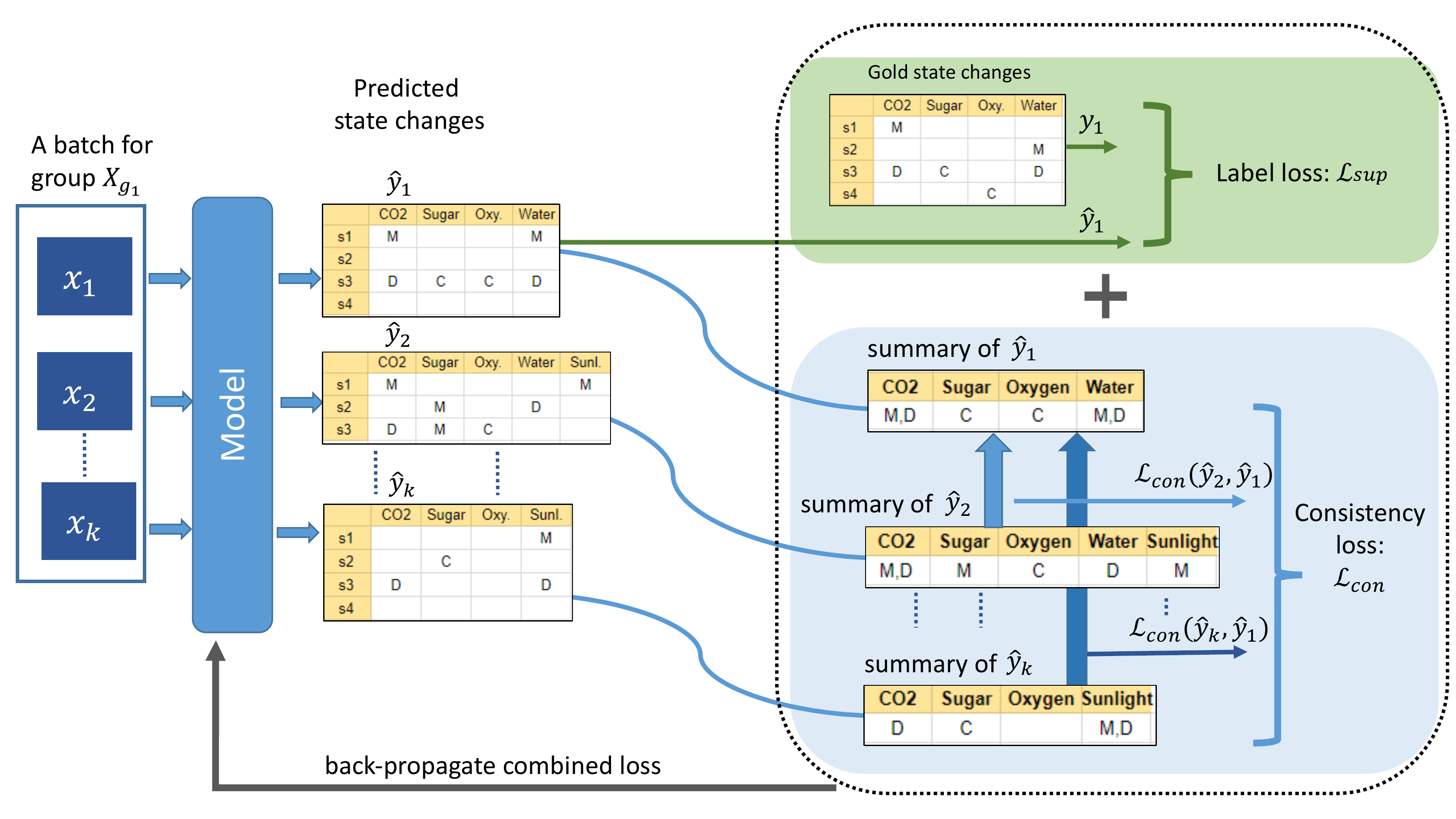}}
\end{center}
\caption{Overview of the \model~training framework, illustrated for the procedural comprehension task ProPara. During training, \model~processes {\it batches} of examples \{$x_1$,...,$x_k$\} \finaledit{for each group $X_{g}$}, where predictions for one example (here $\hat{y_1}$) are compared against its gold  (producing loss $\mathbb{L}_{sup}$), and its summary against summaries of all other examples to encourage consistency of predictions (producing $\mathbb{L}_{con}$), repeating for each example in the batch.}
\label{fig:lace_model}
\end{figure*}

While a traditional supervised learning model operates on individual examples,
\model~operates on {\it batches} of grouped examples $X_{g}$. Given a group $g$
containing $N$ labeled examples $\{x_1,...,x_N\}$ (we drop the $g$ subscript for clarity),
\model~creates $N$ batches, each containing all the examples but with a
different $x_i$ labeled as ``primary'', along with the gold labels $y_i$ for
(only) the primary example. (We informally refer to the primary example as the ``first example''
in each batch). Then for each batch, \model{} jointly optimizes the usual supervised
loss $\mathbb{L}_{sup} (\hat{y_{i}}, y_{i})$ for the primary example, along with a
consistency loss between (summary) predictions for all other members of the group and the
primary example, $\mathbb{L}_{con}(\hat{y_{j}}, \hat{y_{i}})$ for all $j \neq i$.
This is illustrated in Figures~\ref{fig:lace_model} and~\ref{fig:prossl_batching_only}.
This is repeated for all batches.

For example, for the three paragraphs about photosynthesis (Figure~\ref{fig:example_multiple_paragraphs}),
batch 1 compares the first paragraph's predictions with its gold labels, and also
compares the summary predictions of paragraphs 2 and 3 with those of the first paragraph
(Figure~\ref{fig:prossl_batching_only}). This is then repeated using paragraph 2, then paragraph 3 as primary.

The result is that \model{} jointly optimizes
the supervised loss $\mathbb{L}_{sup}$ and consistency loss $\mathbb{L}_{con}$ to
train a model that is both accurate for the given task as well as consistent in its
predictions across examples that belong to the same group.

This process is approximately equivalent to jointly optimizing the usual supervised
loss $\mathbb{L}_{sup} (\hat{y_{i}}, y_{i})$ for all examples in the group,
and the pairwise consistency loss $\mathbb{L}_{con}(\hat{y_{j}}, \hat{y_{i}})$
for all pairs $(x_j,x_i), \finaledit{j \neq i}$ in the group. However, there is an important difference,
namely the relative contributions of $\mathbb{L}_{sup}$ and $\mathbb{L}_{con}$
is varied among batches, depending on how accurate the predictions for the primary
example are (i.e., how small $\mathbb{L}_{sup}$ is), as we describe later in
Section~\ref{sec:consistency}. This has the effect of
paying more attention to consistency loss when predictions on the primary
are more accurate.

We also extend \model~to the semi-supervised setting as follows.
For the semi-supervised setting, where only $m$ of $n$ $(m < n)$ examples are labeled,
we only form $m$ batches, where each batch has a different labeled example as primary.
We later report experiments results for both the fully and semi-supervised settings.

\subsection{Base Model for Procedural Text}

We now describe how \model~is applied to our goal of comprehending procedural text.
Note that \model~is agnostic to the learner used within the framework.
For this application, we use a simplified version of ProStruct \cite{propara-emnlp18},
a publicly available system designed for the ProPara task.
Our implementation simplifies ProStruct by reusing its encoder,
but then predicting (a distribution over) each state change label independently during
decoding for every cell in the $|S| \times |E|$ grid (Figure~\ref{fig:example_multiple_paragraphs}). We briefly summarize this here.

\subsubsection{Encoder}
\label{sec:encoder}
ProStruct uses an encoder-decoder architecture that takes procedural text as input and predicts the state changes of entities $E$ in the text as output. During encoding, each step $s_t$ is encoded using $\vert E \vert$ embeddings, one for each entity $e_j \in E$. Each embedding represents the action that $s_t$ describes, applied to $e_k$. The model thus allows the same action to have different effects on different entities (e.g., a transformation destroys one entity, and creates another).

For each $(s_t,e_j) \in S \times E$ pair, the step is fed into a BiLSTM \cite{hochreiter1997long}, using pretrained GloVe \cite{pennington2014glove} vectors $v_w$ for each word $w_i$
concatenated with two indicator variables, one indicating whether $w_i$ is a word referring to $e_j$, and one indicating whether $w_i$ is a
verb. A bilinear attention layer then computes attention over the  contextualized vectors $h_i$ output by the BiLSTM: $a_i = h_i * B * h_{ev} + b$ , where $B$ and $b$ are learned parameters, and $h_{ev}$ is the concatenation of $h_e$  (the averaged contextualized embedding for the entity words $w_e$) and $h_v$ (the averaged contextualized embedding for the verb words $w_v$).

Finally, the output vector $c_{tj}$ is the attention-weighted sum of the $h_i$: $c_{tj} = \sum_{i=1}^I a_{i} * h_i$ . Here, $c_{tj}$ can be thought of as representing the action $s_t$ applied to entity $e_j$. This is repeated for all steps and entities.

\subsubsection{Decoder}

To decode the action vectors $c_{tj}$ into their resulting state changes they imply, 
each is passed through a feedforward layer to generate $logit(\pi_{tj})$, a set of logistic
activations over the $K$ possible state changes $\pi_{tj}$ for entity $e_j$ in step $s_t$.
For ProPara, there are $K=4$ possible state changes: \statechange{Move, Create, Destroy,} and \statechange{None}.
These logits form a distribution over possible state changes to predict, for
each entity and step in the text. We then compute loss, described next,
using these distributions directly rather than discretizing them into exact
predictions at this stage, so as not to lose information.

\subsection{Applying \model}

\label{sec:consistency}

\subsubsection{Batching}

We start by creating training batches for each $X_{g_i} \in X_{g}$. From a group $X_{g_i}$ comprising of $n$ examples, we create $n$ training batches. A batch consists of all $n$ examples $(x_1, x_2, ..., x_n)$, but the loss computation is different in each batch. Figure \ref{fig:prossl_batching_only} illustrates this.

\subsubsection{Loss Computation}

The loss computation in a batch is based on the usual \finaledit{supervised} loss and additionally the consistency loss, as follows:
\begin{align}
\label{eq:loss}
\mathbb{L}_{batch} &= \lambda \underbrace{\mathbb{L}_{sup}(\hat{y_1}, y_1)}_{supervised~loss} ~+ (1-\lambda) \underbrace{\sum_{i=2}^{n} \mathbb{L}_{con}( \hat{y_i},\hat{y_1})}_{consistency~loss}
\end{align}

Here, $\mathbb{L}_{sup}(\hat{y_1}, y_1)$ is the negative log likelihood loss\footnote{\finaledit{Loss function $\mathbb{L}_{sup}$ is exactly same as the loss function used in the base model so that we can measure the effect of adding consistency loss.}} against the gold labels $y_1$, and $\lambda$ is a hyperparameter tuned on the dev set.

To compute the consistency loss $\mathbb{L}_{con}(\hat{y_i}, \hat{y_1})$, we compare the summaries
computed from $\hat{y_i}$ and $\hat{y_1}$. In our particular application, a summary lists all the state changes each entity undergoes, formed by aggregating its step-by-step state changes. For example, for paragraph $x_1$ in Figure~\ref{fig:lace_model},
as \texttt{CO$_2$} first moves (\texttt{M}), then later is destroyed (\texttt{D}), we summarize its state changes
as $s(\texttt{CO}_2,\hat{y}_1)$ = \{\texttt{M,D}\}. In practice, as our decoder outputs distributions over the four possible
values \{\texttt{M,C,D,N}\} rather than a single value, we summarize by adding and normalizing these
distributions, producing a summary distribution $s(e,\hat{y}_j)$ over the four values
rather than a discrete set of values.

To compute the consistency loss $\mathbb{L}_{con}(\hat{y_i}, \hat{y_1})$ itself,
we compare summaries for each entity $e$ that occurs in both paragraph \finaledit{${x_1}$ and 
paragraph $x_i$ (referred to as Ent($x_1$) and Ent($x_i$) respectively)}, and compute the average mean squared error (MSE) between their summary distributions. \finaledit{We also tried other alternatives (e.g., Kullback-Leibler divergence) for calculating the distance between summary distributions, but mean squared error performs best. Equation~\ref{eq:tau} shows the details for computing the consistency loss.} 

\begin{align}
\label{eq:tau}
\hspace{-2mm} \mathbb{L}_{con}(\hat{y_i}, \hat{y_1}) &= \frac{\sum_{e \in Ent(x_i) \cap Ent(x_1)} \textrm{MSE}(s(e, \hat{y}_i),s(e, \hat{y}_1))}{\vert Ent(x_i) \cap Ent(x_1) \vert}
\end{align}

\finaledit{
Note that each paragraph contains varying number of entities and sentences. It is possible that some paragraphs do not mention  exactly the same entities
as the labeled paragraph (first element in the batch). In such cases, we penalize the model only for predictions for co-occurring entities. Unmatched entities are not penalized.
}
\subsubsection{Adaptive Loss}
The supervised loss $\mathbb{L}_{sup}(\hat{y_1}, y_1)$ is large in the early epochs when the model is not sufficiently trained. At this point, it is beneficial for the model to pay no attention to the consistency loss $\mathbb{L}_{con}(\hat{y_j}, \hat{y_1})$ as the predicted action distributions are inaccurate. To implement this, if $\mathbb{L}_{sup}$ is above a defined threshold then the consistency loss term in Equation~\ref{eq:loss} is ignored (i.e. $\lambda = 1$). Otherwise, Equation~\ref{eq:loss} is used as is. 
\finaledit{This can loosely be seen as a form of simulated annealing \cite{simulated_annealing_1988}, using just two temperatures. Note that the time (epoch number) when the temperature (lambda) changes will vary across batches depending on the supervised loss within that batch of data, hence we call it an ``adaptive'' loss.}

\section{Experimental Results}
We now present results on ProPara, the procedural text comprehension dataset introduced in \cite{propara-naacl18}. There are 187 topics in this dataset and a total of 488 labeled paragraphs (around 3 labeled paragraphs per topic). 
The task is to track how entities change state through the paragraph (as described in Section~\ref{instantiation})
and answer 4 classes of questions about those changes (7043/913/1095 questions in each of the train/dev/test partitions respectively).
We compare \model{} with the baselines and prior state-of-the-art model ProStruct \cite{propara-emnlp18} in two settings: (1) Fully supervised learning (using all the training data). (2) Semi-supervised learning (using some or all of the training data, plus additional unlabeled data).


\subsection{Fully Supervised Learning}

\begin{table}[tbh]
\resizebox{\columnwidth}{!}{%
\begin{tabular}{l|ccc}
\toprule 
Models                & P    & R    & F1   \\ \midrule
EntNet~\cite{Henaff2016TrackingTW}  &  54.7  &  30.7  &  39.4 \\
QRN~\cite{Seo2017QueryReductionNF}  &  60.9  &  31.1  &  41.1 \\
ProLocal~\cite{propara-naacl18}  &  \textbf{81.7}  &  36.8  &  50.7 \\
ProGlobal~\cite{propara-naacl18}  &  61.7  &  44.8  &  51.9 \\
\prostruct{}~\cite{propara-emnlp18} &  74.3  &  43.0  &  54.5 \\ 
\midrule
\proSSL{}  (our model)   & 75.3 &\textbf{45.4}	& \textbf{56.6}    \\
\bottomrule
\end{tabular}}
\caption{Comparing the performance of \proSSL~with prior methods on the test partition of ProPara.}
\label{table:prossl_on_emnlp18}
\end{table}

We evaluated \model~by comparing its performance against published, state-of-the-art results
on ProPara, using the full training set to train \model. The results
are shown in Table~\ref{table:prossl_on_emnlp18}.
In Table \ref{table:prossl_on_emnlp18}, all the baseline numbers are the results
reported in \cite{propara-emnlp18}. Note that all these baselines are trying to reduce the gap between predicted labels and gold labels on the training dataset. \model{}, however,
also optimizes for consistency across labels for 
{\it groups} of paragraphs belonging to the same topic. As \model~uses parts of ProStruct   as its learning algorithm, the gains over ProStruct appear to be coming directly from its novel learning framework described in Section \ref{sec:lace_overview}. To confirm this, we also performed an ablation study,
removing the consistency loss term and just using the base model in \model.
The results are shown in Table \ref{table:prossl_ablation}, and show that
the F1 score drops from 56.6 to 53.2, illustrating that the consistency loss
is responsible for the improvement. In addition,
Table \ref{table:prossl_ablation}~indicates that consistency loss helps
improve both precision and recall. 

Also note that \model{}~simplifies parts of ProStruct. For example, unlike ProStruct, \model~does not use a pre-computed knowledge base during decoding. Thus \model{} is more efficient to train than ProStruct (>15x faster at training time). 

Finally, \model{} builds upon  ProStruct (state-of-the-art when we started working on our model).
  Since \model{} was developed, two higher results of 57.6 and 62.5 on the ProPara task have appeared  \cite{Das2018BuildingDK,Gupta2019TrackingDA}. Both systems are fully supervised and developed contemporaneously with \model.
  In principle \model's approach of leveraging consistency across paragraphs to train a more robust model can be applied to other systems. Our main contribution
  is to show that maximizing consistency across datapoints (in addition to minimizing supervised loss) enables a model to leverage unlabeled data and leads to more robust results.
  
\begin{table}[tb]
\centering
\begin{tabular}{l|ccc}
\toprule 
Models                & P    & R    & F1   \\ 
\midrule
\proSSL               & \textbf{75.3}    & \textbf{45.4}	& \textbf{56.6 }      \\ 
- consistency loss $\mathbb{L}_{con}$    &  69.6	& 43.1  & 53.2\\

\bottomrule
\end{tabular}
\caption{\proSSL~ablation results}
\label{table:prossl_ablation}
\end{table}

\eat{
\begin{table}[!tb]
\begin{tabular}{lllll}
\\\toprule
& Train & Dev  & Test & Total \\ \midrule
\#  topics                                                    & 147   & 18   & 18  & 488 \\
\# labeled paragraphs                                                & 391   & 43   & 54   \\ \midrule
\begin{tabular}{@{}l@{}} \# unlabeled paragraphs \normalsize \end{tabular}      & 437   & 49   & 61  & 877 \\
\bottomrule
\end{tabular}
\caption{Statistics of the unlabeled data aligned with ProPara topics.}
\label{table:stats}
\end{table}
}

\begin{table}[t]
\centering
\resizebox{\columnwidth}{!}{
\begin{tabular}{l|ccc}
\toprule
Models   & \multicolumn{3}{c}{Proportion of labeled paragraphs } \\ 
   & \multicolumn{3}{c}{used per training topic} \\ \midrule
          & 33\%       & \ \ \ \ \ \ \ 66\%         & 100\%       \\ \midrule
\prostruct{} &  45.4  
          &  \ \ \ \ \ \ \ 50.6  
          &  54.5  
          \\ \midrule
\proSSL &   47.3    
       &   \ \ \ \ \ \ \ 51.2    
       &   56.6   \\  
\proSSL~+ unlabeled data  &  49.9 
                  &  \ \ \ \ \ \ \ 52.9 
                  &  56.7 
                  \\ \bottomrule
\end{tabular}
}
\caption{Comparing \proSSL~vs. \prostruct{} with varying amount of labeled paragraphs available per training topic. We compare their performance in terms of F1 on ProPara test partition.}
\label{table:vary_training_data_1}
\end{table}


\subsection{Semi-Supervised Learning}

Unlike the other systems in Table~\ref{table:prossl_on_emnlp18},
\model~is able to use unlabeled data during training. 
As described in Section \ref{sec:lace_overview}, given a group containing both labeled and unlabeled paragraphs, we create as many batches as the number of labeled paragraphs in the group. Hence, paragraphs $x_i$ with gold labels $y_i$ can contribute to both supervised loss $\mathbb{L}_{sup}$ and 
consistency loss $\mathbb{L}_{con}$. Additionally, we can use unlabeled paragraphs $x_j$ (i.e., without gold labels $y_j$), while computing consistency loss $\mathbb{L}_{con}$. This way \model{} can make use of unlabeled data during training. 


To evaluate this, we collected 877 additional unlabeled paragraphs for ProPara topics\footnote{\finaledit{The unlabeled paragraphs are available at \url{http://data.allenai.org/propara/}.}}.
\finaledit{
As the original ProPara dataset makes some simplifying assumptions, in particular that events are mentioned in chronological order,
we used Mechanical Turk to collect additional paragraphs that conformed to those assumptions (rather than collecting paragraphs from Wikipedia, say).
Approximately 3 extra paragraphs were collected for each topic in ProPara.}
\finaledit{Note that collecting unlabeled paragraphs is substantially less expensive than labeling paragraphs.} 
 
\begin{table}[!htb]
\centering
\begin{tabular}{ccccc}
\toprule 
                & Train    & Dev    & Test & Unlabeled   \\ 
\midrule
 \# paragraphs & 391    & 54	& 43  &  877   \\ 
\bottomrule
\end{tabular}
\caption{\finaledit{ProPara Paragraphs Statistics}}
\label{table:unlabeled_stats}
\end{table}



We then trained the \prostruct~and \model~models varying two different parameters:
(1) the percentage of the labeled (ProPara) training data used to
train the system (2) for \model~only, whether the additional unlabeled
data was also used. This allows us to see performance under different
conditions of sparsity of labeled data, and (for \model) also assess how much unlabeled
data can help under those conditions.
During training, the unused labeled data was ignored (not used as unlabeled data).
We keep the dev and test partitions the same as original dataset, picking a model based on dev performance and report results on test partition. The results are 
shown in Table \ref{table:vary_training_data_1}. In the first two rows,
\prostruct{} and \model{} are both trained with x\% of labeled data, while the last row reports performance of \model{} when it also has access to new unlabeled paragraphs.

\begin{figure}[t]
\begin{center}
\includegraphics[width=\columnwidth]{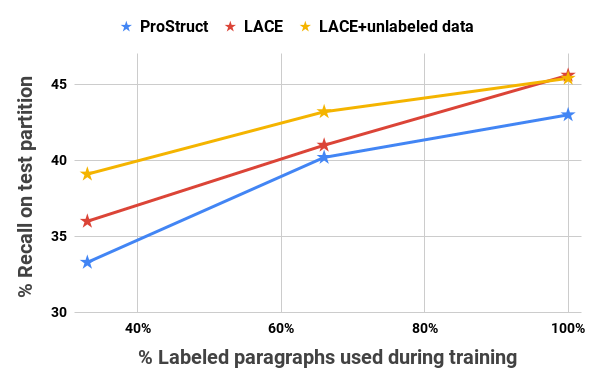}
\end{center}
\caption{Comparing  \proSSL~vs. \prostruct{} based on Recall on the test partition, by varying amount of labeled paragraphs available per training topic}
\label{fig:recall_vary_labeled_data}
\end{figure}

Table \ref{table:vary_training_data_1} demonstrates that \model{} results in even larger improvements over \prostruct{} when the amount of labeled data is limited. In addition, unlabeled data adds an additional boost to this performance, in particular when
labeled data is sparse. Further examination suggests that the gains in F1 are resulting mainly from improved recall, as shown in Figure \ref{fig:recall_vary_labeled_data}. We believe that having access to unlabeled paragraphs and optimizing consistency across paragraphs for training topics, helps \model{} generalize better to unseen topics.

\subsection{\finaledit{Implementation Details for \model}}

We implement our proposed model \model{} in PyTorch \cite{pytorch} using the AllenNLP \cite{allennlp} toolkit. We added a new data iterator that creates multiple batches per topic (Figure \ref{fig:prossl_batching_only}) which enables easy computation of consistency loss. 
We use 100D Glove embeddings \cite{pennington2014glove}, trained on Wikipedia 2014 and Gigaword 5 corpora (6B tokens, 400K vocab, uncased). Starting from glove embeddings appended by entity and verb indicators, we use bidirectional LSTM layer to create contextual representation for every word in a sentence. We use 100D hidden representations for the bidirectional LSTM~\cite{hochreiter1997long} shared between all inputs (each direction uses 50D hidden vectors). We use attention layer on top of BiLSTM, using a bilinear similarity function similar to~\cite{Chen2016ATE} to compute attention weights over the contextual embedding for each word in the sentence. 

To compute the likelihood of all state changes individually, we use a single layer feedforward network with input dimension of 100 and output 4. In these experiments, we check if the supervised loss $\mathbb{L}_{sup}$ is less than a threshold (0.2 in our case) then we use equation \ref{eq:loss} and $lambda=0.05$. All hyper-parameters are tuned on the dev data.

During training we use multiple paragraphs for a topic to optimize for both supervised and consistency loss. At test time, \model's predictions are based on only one given paragraph. All the performance gains are due to the base model being more robust due to proposed training procedure. The code for \model{} model is published at \url{https://github.com/allenai/propara}.

\subsection{ Analysis and Discussion}
\label{sec:lace_analysis}
We first discuss the predicted label consistency across paragraphs for \model{} vs. \prostruct{}. We then identify some of the limitations of \model{}. 

\subsubsection*{Label Consistency}

\begin{table}[t]
\centering
\begin{tabular}{l|cc}
\toprule
          & \multicolumn{2}{c}{Consistency Score (\%)}                \\ \midrule
          & \multicolumn{1}{c}{Train} & \multicolumn{1}{c}{Test} \\ \midrule
ProStruct & 46.70                     & 37.21                    \\
\model{}      & 54.39 &  38.36                   \\
\bottomrule
\end{tabular}
\caption{Consistency score comparison}
\label{table:consis}
\end{table}

\model~attempts to encourage consistency between paragraphs about the same 
topic during training, and yield similar benefit at test time.
To examine whether this happens in practice, we compute and 
report the consistency score between paragraphs about the same topic (Table~\ref{table:consis}). Specifically, for an entity that appears in two paragraphs 
about the same topic, we compare whether the summaries of state change predictions for each match. The results are shown in Table~\ref{table:consis}. 

The table shows that \model{} achieves greater prediction consistency during training, and that this benefit plays out to 
some extent at test time even though label consistency is not enforced at test time
\finaledit{
(we do not assume that examples are grouped at test time, hence consistency
between groups cannot be enforced as the grouping is unknown).
}
As an illustration, for the topic \emph{describe the life cycle of a tree} which is unseen at training time, for the three paragraphs on the topic, ProStruct predicts that tree is created; not-changed; and created respectively, while \model{} correctly predicts that tree is created; created; and created respectively. This illustrates a case where \model{} has learned to make predictions that are more consistent and correct. 

\eat{
Table~\ref{table:consis} shows that this improvement generally holds,
although the improvement on test is smaller than on train. 
Our hypothesis is that \model{} is more optimized towards topics in train which are unseen at test time.
}
\eat{
\model{} achieves greater prediction consistency during training.
For example, for the three paragraphs on the topic ``How do plants obtain and use water?'', \prostruct{} predicts that water is not-changed; not-changed
and moved respectively, while \model{} correctly predicts that water is moved;
moved; and moved respectively. This illustrates a case where \model{}
has learned to make predictions that are more consistent and correct across paragraphs.
\todo{Find a dev set example here} Table~\ref{table:consis} shows
that this improvement generally holds. But as compared to the larger consistency score improvement on train, the improvement on test is smaller. Our hypothesis is that \model{} is more optimized towards topics in train which are unseen at test time.

Leveraging label consistency at training time makes \model{} more robust compared to ProStruct.
We think that this is the reason behind performance improvements  in Table \ref{table:prossl_on_emnlp18} 
 and Table \ref{table:vary_training_data_1},  even though label consistency is not used at test time.
 }

\subsubsection*{Error Analysis for \model{}}
To understand \model's behavior further, we examined cases where
\model's and \prostruct's predictions differ, and examined their agreement with gold labels. In this analysis we found three major sources of errors for \model:

\begin{itemize}[leftmargin=*]
    \item \textbf{ The label consistency assumption does not always hold:} In Section \ref{sec:problem-definition}, we explain that \model{} relies on summary labels being consistent across examples in the same group. We found that for some of the topics in our training dataset this assumption is sometimes violated.
    E.g., for the topic  \emph{How does the body control its blood sugar level?}, there are two different paragraphs; one of them describes the entity \entity{\entity{sugar}} as being \underline{\entity{Created}} and then \underline{\entity{Destroyed}} to create \entity{bloodsugar}, while the other 
    paragraph describes the same event in a different way by saying that the entity \entity{sugar} is \underline{\statechange{Created}} and then \underline{\statechange{Moved}} to the blood. 
    \model{} can thus goes wrong when trying to enforce consistency in such cases. 
    
    \item \textbf{Lexical variance between entities across paragraphs:} Different paragraphs about the same topic may describe the procedure using different wordings,
    resulting in errors. For example, in paragraphs about the topic \emph{what happens during photosynthesis?}, the same entity (\entity{carbon dioxide}) is referred to by two different strings, \entity{CO$_2$} in one paragraph and \entity{carbon dioxide} in another. Currently, \model{} does not take into account entity synonyms, so it is unable to encourage consistency here. \finaledit{An interesting line of future work would be to use the embedding space similarity between entity names, to help address this problem.}
    
    \item \textbf{\model{} can make incorrect predictions to improve consistency:} For the topic \emph{Describe how to make a cake} at training time, when presented with two paragraphs, \model{} tries to be consistent and incorrectly predicts that \entity{cake} is \statechange{Destroyed} in both paragraphs.  ProStruct does not attempt to
    improve prediction consistency, here resulting in less consistent but in this case more accurate predictions for this topic.
\end{itemize}

\subsection{\finaledit{Directions For Enhancing \model}}
\begin{itemize}
    \item \finaledit{Improve \model{} for ProPara:}
    \model's performance on ProPara can be improved further by a) soft matching of entities across paragraphs instead of current exact string match b) exploring more systematic ways (e.g., simulated annealing) to define adaptive loss c) using additional sources of unlabeled data (e.g., web, textbooks) weighed by their reliability.   
    
    \item \finaledit{Apply \model{}  on other tasks:} 
    Architecturally, \model{} is a way to train any existing structured prediction model for a given task to produce consistent labels across similar datapoints.
    Hence it can be easily applied to other tasks where parallel data is available (grouping function) and there is a way to efficiently compare predictions (summary labels) across parallel datapoints, e.g. event extraction from parallel news articles \cite{Chinchor2002MessageUC}. 
    
    Further, summary labels need not be action categories (e.g., \underline{\entity{Created}}, \underline{\entity{Destroyed}}). Consistency can also be computed for QA task where multiple parallel text is available for reading comprehension. We plan to explore this direction in the future. 
\end{itemize}

\section{Conclusion}

Our goal is procedural text comprehension, a task that current
systems still struggle with. Our approach has been to exploit
the fact that, for many procedures, multiple independent descriptions
exist, and that we expect some consistency between those descriptions.
To do this, we have presented a task- and model-general learning
framework, \model, that can leverage this expectation, allowing consistency
bias to be built into the learned model. Applying this framework
to procedural text, the resulting system obtains new state-of-the-art
results on the ProPara dataset, an existing benchmark for procedural
text comprehension. It also demonstrates the ability to benefit
from unlabeled paragraphs (semi-supervised learning), something that
prior systems for this task were unable to do. We have also
identified several avenues for further improvement (Section~\ref{sec:lace_analysis}),
and are optimistic that further gains can be achieved.

\subsubsection*{Acknowledgements}
Computations on \href{https://beaker.org}{beaker.org} were supported in part by credits from Google Cloud.

\bibliography{references}

\end{document}